\documentclass[letterpaper, 10 pt, conference]{ieeeconf}  

\IEEEoverridecommandlockouts                              

\usepackage{graphics}
\usepackage{graphicx} 
\usepackage{epsfig} 
\usepackage{hyperref}

\hypersetup{hypertex=true,
            colorlinks=true,
            linkcolor=blue,
            anchorcolor=blue,
            citecolor=blue}
\usepackage{booktabs}
\usepackage{todonotes}
\usepackage{float} 
\usepackage{multicol} 
\usepackage{subcaption}
\usepackage{authblk} 
\usepackage{multirow}
\usepackage{cite}
\usepackage[T1]{fontenc}
\usepackage{caption}
\title{\LARGE \bf
Generalized Robot Learning Framework
}

\author{
  Jiahuan Yan$^{1}$, Zhouyang Hong$^{1}$, Yu Zhao$^{1}$, Yu Tian$^{2}$, Yunxin Liu$^{3}$, Travis Davies$^{1}$, Luhui Hu$^{1}$ \\
  $^{1}$ZhiCheng AI, $^{2}$Harvard University, $^{3}$Tsinghua University
}

\begin{document}
\maketitle
\thispagestyle{empty}
\pagestyle{empty}



\begin{abstract}
Imitation based robot learning has recently gained significant attention in the robotics field due to its theoretical potential for transferability and generalizability. However, it remains notoriously costly, both in terms of hardware and data collection, and deploying it in real-world environments demands meticulous setup of robots and precise experimental conditions. In this paper, we present a low-cost robot learning framework that is both easily reproducible and transferable to various robots and environments. We demonstrate that deployable imitation learning can be successfully applied even to industrial-grade robots, not just expensive collaborative robotic arms. Furthermore, our results show that multi-task robot learning is achievable with simple network architectures and fewer demonstrations than previously thought necessary. As the current evaluating method is almost subjective when it comes to real-world manipulation tasks, we propose Voting Positive Rate (VPR)—a novel evaluation strategy that provides a more objective assessment of performance. We conduct an extensive comparison of success rates across various self-designed tasks to validate our approach. To foster collaboration and support the robot learning community, we have open-sourced all relevant datasets and model checkpoints, available at \href{https://huggingface.co/ZhiChengAI}{https://huggingface.co/ZhiChengAI}
\end{abstract}

\section{Introduction}
Robotic research \cite{c1,c2,c45} has increasingly focused on the potential of using imitation learning \cite{c48} for robotic manipulation \cite{c49}, driven in part by the growing integration of generative AI \cite{c50} into industry. However, the high cost of current robot learning pipelines has remained a significant barrier, limiting access to practical development and scaling up. Hence, we propose a low-cost real robot imitation learning framework that facilitates efficient data collection, training, and inference on a industrial-grade robotic arm with common household control devices, making it possible for a broader range of researchers and practitioners to engage in robotics innovation.


To rigorously evaluate the effectiveness of the proposed framework, we designed 10 distinct robotic tasks, each characterized by specific features tailored to real-world conditions. A comprehensive analysis of these tasks is provided, encompassing both the rationale behind their design and their empirical performance in deployment. Section \ref{sec:Task Design} delves into the requirements and methodologies involved in the design of these tasks, while Section \ref{sec:Experiments and Results} systematically examines how task-specific characteristics impact performance outcomes in real-world testing scenarios.

In addition to the versatility of using a general-purpose robotic arm that can meet the demands of various industrial scenarios, our framework also demonstrates model generalization. Specifically, we successfully enable a single checkpoint to perform multiple tasks by combining datasets and applying minor adjustments to the training strategy, which are discussed in Section \ref{sec:Dataset Scaling Matters} and Section \ref{sec:Model Generalization}.
\begin{figure}
    \centering
    \includegraphics[width=\linewidth]{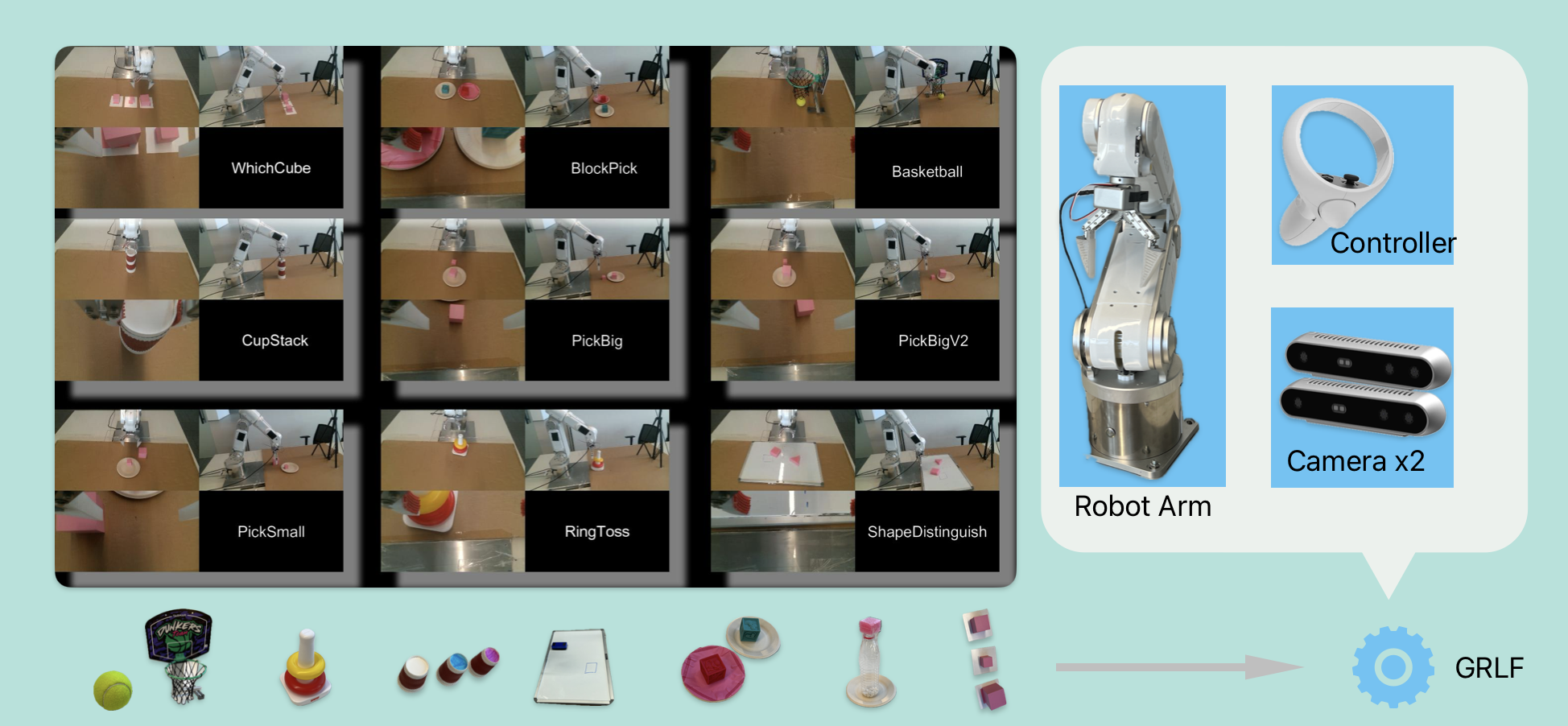}
    \caption{Overview of the framework: A real-world robot learning setup can be constructed using everyday household items, a robotic arm, a controller and two cameras.}
    \label{fig:enter-label}
\end{figure}
To support the broader research community, we also release the entire dataset generated through our framework. This dataset, encompassing diverse tasks and environmental conditions, serves as an additional resource for future research in robot learning, promoting reproducibility and enabling further advancements in the field.

Our main contributions can be summarized as:
\begin{itemize}
    \item We introduce a novel low-cost imitation learning framework that is accessible even to individual researchers.
    \item We collected over 4,000 episodes across 10 distinct real-world robotic tasks, which are publicly released alongside our findings on the correlation between task difficulty and performance.
    \item We demonstrate model generalization by successfully enabling task adaptation through minimal dataset integration and slight modifications to the training process.
\end{itemize}

\begin{figure*}[ht]
    \centering
    \includegraphics[width=\textwidth]{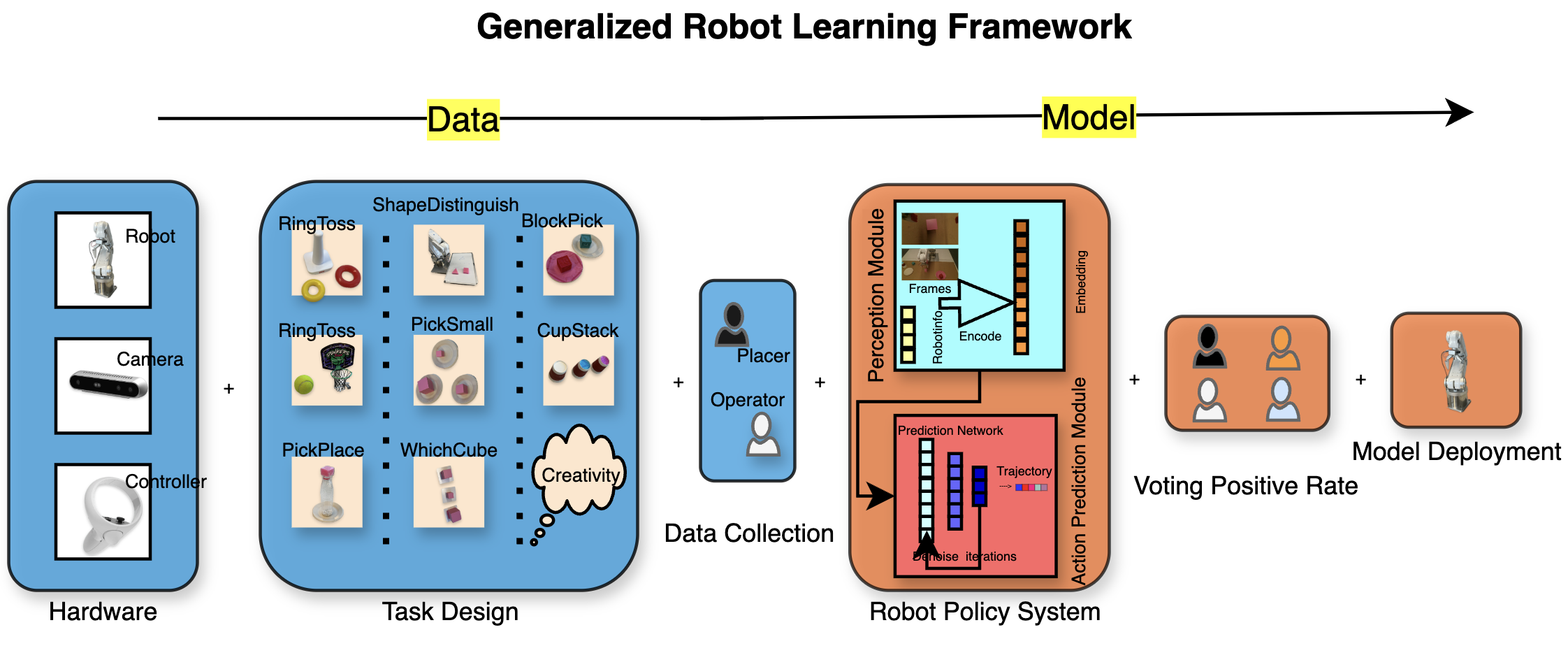}
    \caption{End-to-End Framework: The pipeline illustrates the end-to-end process for a cost-efficient imitation learning implementation, from hardware setup and task design to data collection, modeling and training, evaluation (Voting Positive Rate), and model deployment. This framework is designed to be structurally simple and economically feasible for deployment.}
    \label{fig:framework}
\end{figure*}

\section{Related Work}
\subsection{Imitation Learning} 
Imitation Learning (IL) \cite{c5,c6} is a prominent approach in robotics and autonomous systems, enabling agents to acquire complex behaviors by mimicking expert demonstrations \cite{c7,c8,c9}. Among the various IL techniques, behavioral cloning (BC) has been widely used, framing the task as a supervised learning problem where actions are directly mapped from perceptions \cite{c25,c26}. A well-known example is the ACT policy \cite{c13}, which exemplifies explicit policy learning. However, explicit policies often struggle with multimodal behavior \cite{c24}, as they tend to perform poorly in scenarios with diverse demonstrated actions.

To address these limitations, recent work has explored implicit policies, such as utilizing energy-based models to map perceptions to actions \cite{c24,c27}. While these models offer improved performance in handling multimodal behavior, they present challenges in training stability, particularly due to the requirement of generating negative samples for the Info-NCE loss \cite{c24}.

Diffusion-based policies \cite{c1, c51} addresses the instability of implicit policies by modeling the gradient field of an implicit action score, eliminating the need for negative sampling and improving training stability. Similar to Diffusion Policy, we adopt Denoising Diffusion Probabilistic Models (DDPMs) \cite{c52} as a paradigm for action prediciton. It is well-satisfied with our demand of switching different model structures based on our exhaustive experiment.

\subsection{End-to-End Robot Learning} 
A significant challenge in robotic learning is the availability of diverse and abundant training data. Sim-to-real transfer has been proposed as a solution, where robot models are initially trained in simulation and then adapted to real-world tasks using domain adaptation techniques \cite{c28, c29, c31}. Advances in simulation software, particularly in realistic graphics and physics \cite{c32, c33, c34}, have improved the fidelity of simulated data, making it more comparable to real-world data. However, sim-to-real approaches often fail to capture unforeseen real-world events \cite{c30}, and require specialized skills for setting up and tuning simulations.

Alternatively, end-to-end robot learning offers more accessible, data-driven approach to robot training with imitation learning, as it eliminates many of the complexities of sim-to-real techniques. Projects such as Dobb-E \cite{c35}, ALOHA \cite{c13}, and UMI \cite{c36} have demonstrated successful learning from human demonstrations for end-to-end robot learning. However, these methods are either constrained by professional setups (ALOHA, UMI) or do not involve industrial-grade robotic arms (Dobb-E). In this paper, we propose an end-to-end robot learning framework designed for beginners, capable of operating in less-than-ideal conditions, making robot learning more accessible to non-experts.


\section{Framework Setup}

\subsection{Hardware Preparation}
Our hardware devices for data collection and model deployment facilities are listed as follows:
\begin{itemize}
    \item \textbf{Robotic Arm:} The robotic arm used in our experiments is an industrial-grade model, with a custom-built software development kit (SDK) and a cable connection for communication. While we refrain from disclosing the specific robot model, our framework is designed to be agnostic to the hardware platform. Thus, if our framework is deployable on this robot, it can be adapted for use on almost any robotic system.

    \item \textbf{Cameras:} Two Intel RealSense D415 RGB-D cameras were utilized for frame acquisition. One camera was mounted on the end-effector of the robotic arm to provide a close-up perspective, while the second was positioned to offer a global view of the workspace. The camera type is not a strict requirement, any RGB camera can be used in place of the RealSense D415, depending on individual circumstances.

    \item \textbf{Controller:} An Oculus Quest 2 right-hand controller was employed for data collection. Although the controller facilitated the robot's movement, the headset was also required to ensure a stable spatial coordinate system. The controller's "B" button was programmed to stop the robot's motion, enabling the operator to reposition the controller without interfering with the arm’s operations.
\end{itemize}

This hardware configuration is not rigidly fixed and can be adapted by replacing components based on individual requirements. However, based on our perceptions, it is critical that the hardware used for both data collection and model deployment remain consistent to maintain alignment across the system.



\subsection{Data Collection}
Before arranging human manipulators to collect data for real-task, we took precautions to ensure the controller operated within an obstacle-free area to avoid introducing sudden errors in the control data. To synchronize the operator’s movements with the robotic arm’s movements in real space, we adjusted the orientation of the headset. This adjustment aligned the coordinate systems, providing an intuitive interaction experience where the controller’s movements corresponded directly with the robotic arm’s direction.

We employ a widely adopted strategy for robot data collection, where the trajectory of the robot is recorded alongside timestamps and corresponding video footage. The trajectory data includes the absolute position and orientation of the robot’s end effector (x, y, z, ox, oy, oz), along with additional information indicating the gripper’s state. In our setup, the gripper’s state is represented using Pulse Width Modulation (PWM), which reflects the motor-driven force applied to the gripper.

During data collection, two operators are involved: one arranges the objects based on specific scenarios and their own intuition, while the other uses a controller to remotely operate the robot arm to complete the task. These operators are excluded from the human evaluation process, as discussed in Section \ref{sec:Voting Positive Rate}. The collected data is tagged with the operators’ names to distinguish between identical tasks performed by different collectors, a distinction that serves an important purpose, which will be elaborated on in Section \ref{sec:Data quality}.

In our setup, the number of episodes for each task is not fixed; it is primarily determined by the complexity of the task, including factors such as logical intricacy, the number of objects involved, and the required number of generalization scenarios. As a general guideline, we use 10 episodes per generalization scenario, typically requiring around 100 demonstrations for a specific task. Under optimal conditions, this process takes approximately 0.5 to 1 hour of manual effort. The detailed number is listed in Table \ref{tab:task_performance}.



\subsection{Robot Policy System}
\label{sec:Policy Control System}

Following the design of Diffusion Policy \cite{c1}, we decouple the policy for robot control system into two components: the perception module, which processes information from the physical world and generates embeddings, and the action prediction module, which takes these embeddings as input and outputs corresponding trajectories.
\subsubsection{Perception Module}
Our perception data consists of visual inputs from two cameras: one mounted on the robotic arm’s wrist for a first-person view, and the other providing a third-person perspective. Additionally, low-dimensional state information such as the position and orientation of the end effector and the gripper’s PWM signal is included. This data represents the robot’s perception of the physical world and is processed by a deep learning network to be transformed into embeddings. We experimented with several network architectures, including ResNet18, ResNet34, and ResNet34 configured as a Feature Pyramid Network (FPN) \cite{c40, c41}. Of these, the FPN ResNet34 yielded the best performance, significantly enhancing visual feature representation by leveraging multiple resolutions.
\subsubsection{Action Prediction Module}
This model is designed to map the encoded perception data into a real trajectory that can effectively control the robot arm and gripper to complete tasks. In our current setup, we use the Denoising Diffusion Probabilistic Model (DDPM) \cite{c52} as the backbone, iteratively denoising initially randomized trajectories to make them practical for robot manipulation. We explored two main network architectures within the DDPM framework: Convolutional Neural Networks (CNN) and Transformers.

These two modules function together as a policy control system within our framework, but the choice of network architecture for each module is flexible and not strictly constrained to any specific design.

\subsection{Task Design}
\label{sec:Task Design}
\begin{figure*}[t]
\centering

\begin{minipage}[b]{0.18\textwidth}
    \centering
    \includegraphics[width=\textwidth]{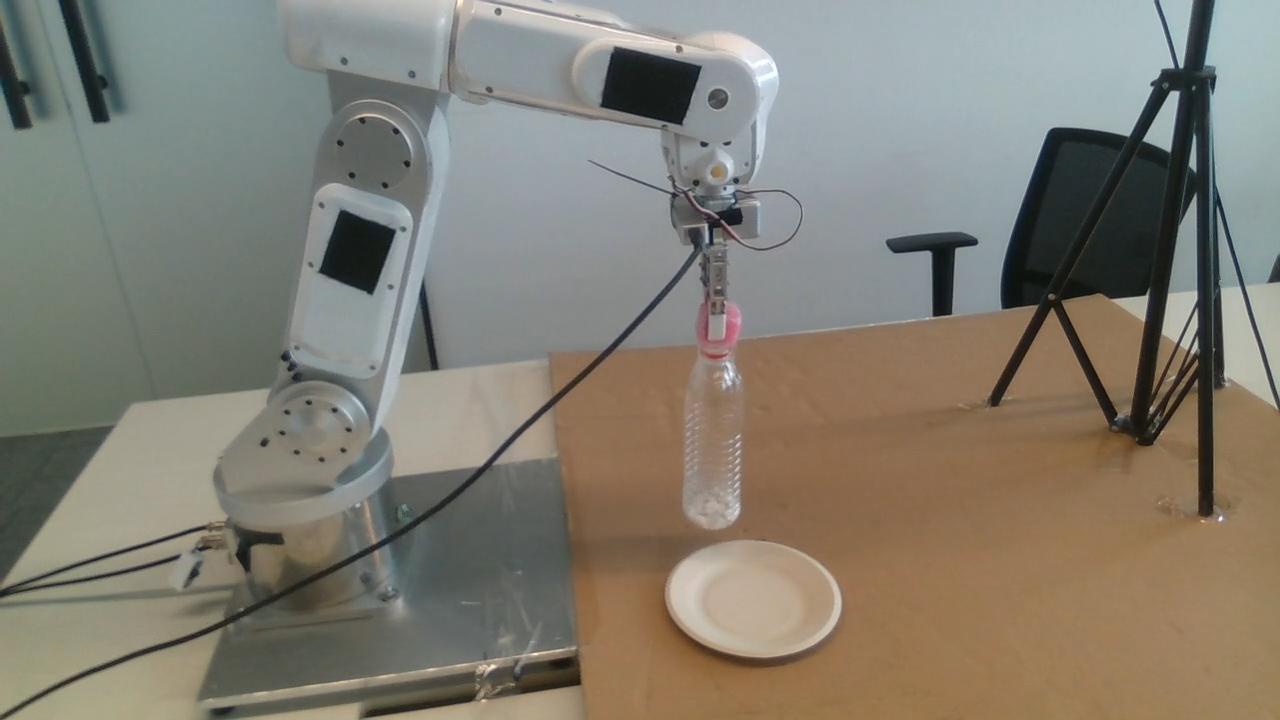}
    \caption*{(a) PickPlace}
\end{minipage}
\hfill
\begin{minipage}[b]{0.18\textwidth}
    \centering
    \includegraphics[width=\textwidth]{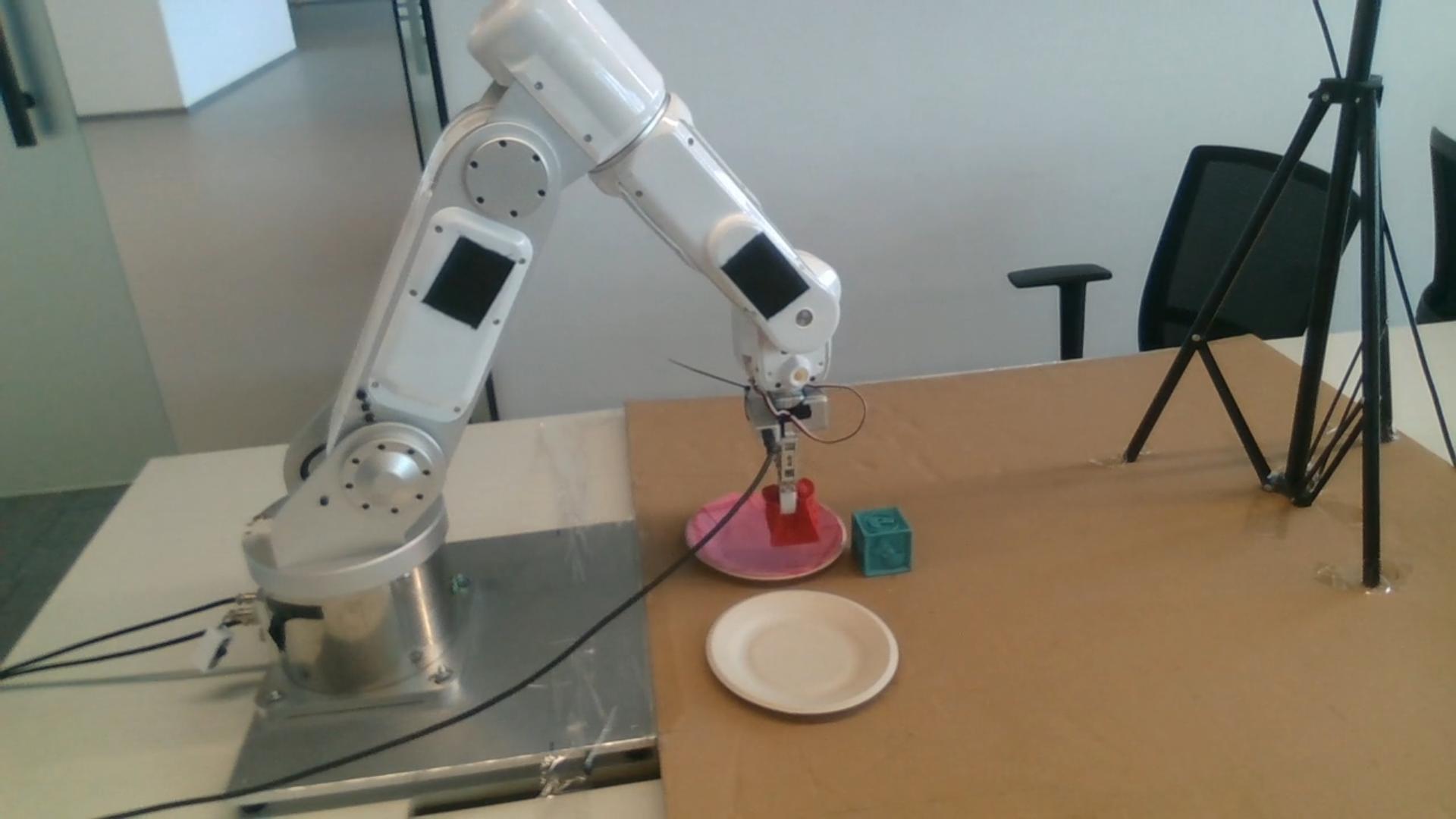}
    \caption*{(b) BlockPick}
\end{minipage}
\hfill
\begin{minipage}[b]{0.18\textwidth}
    \centering
    \includegraphics[width=\textwidth]{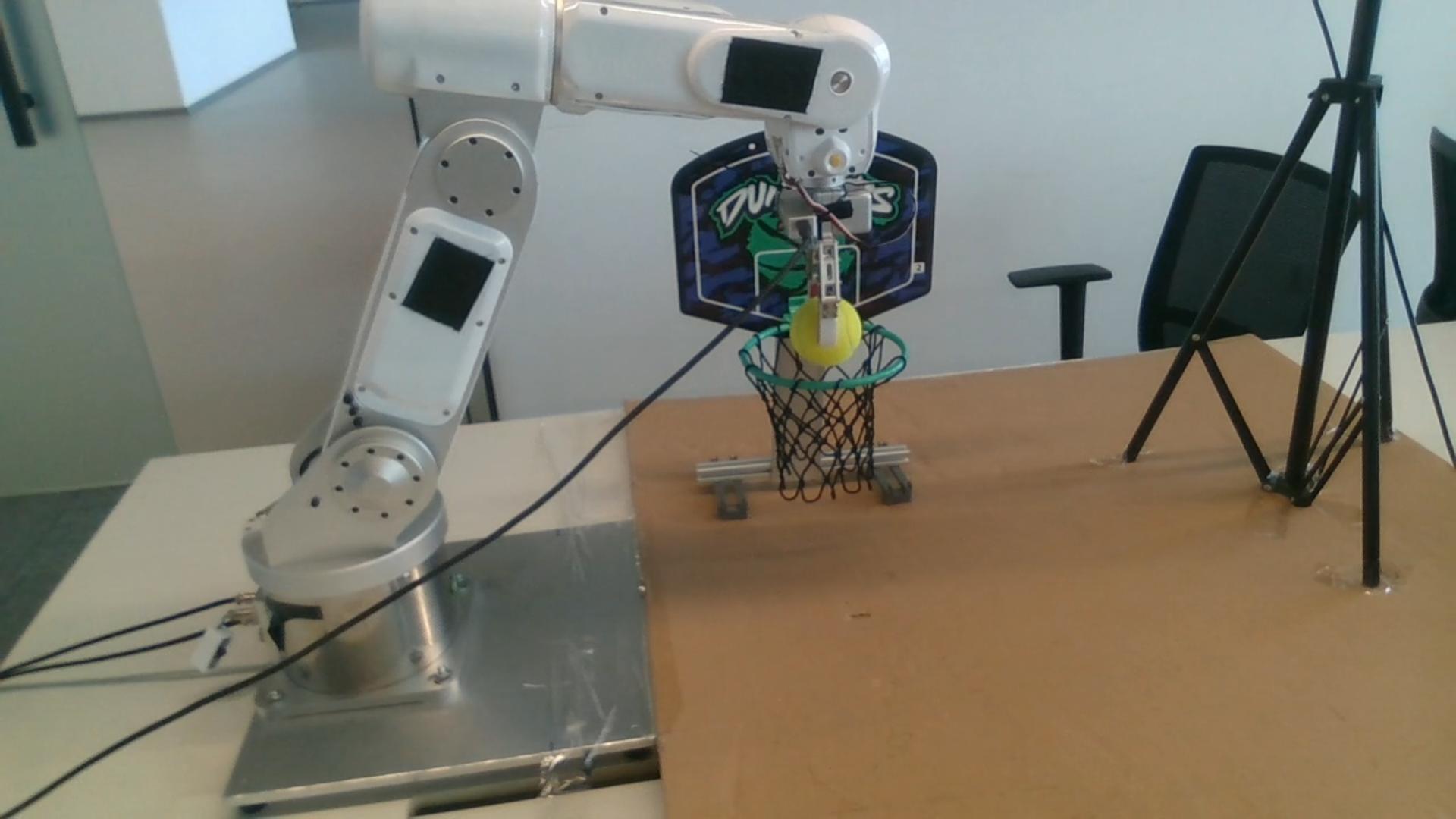}
    \caption*{(c) Basketball}
\end{minipage}
\hfill
\begin{minipage}[b]{0.18\textwidth}
    \centering
    \includegraphics[width=\textwidth]{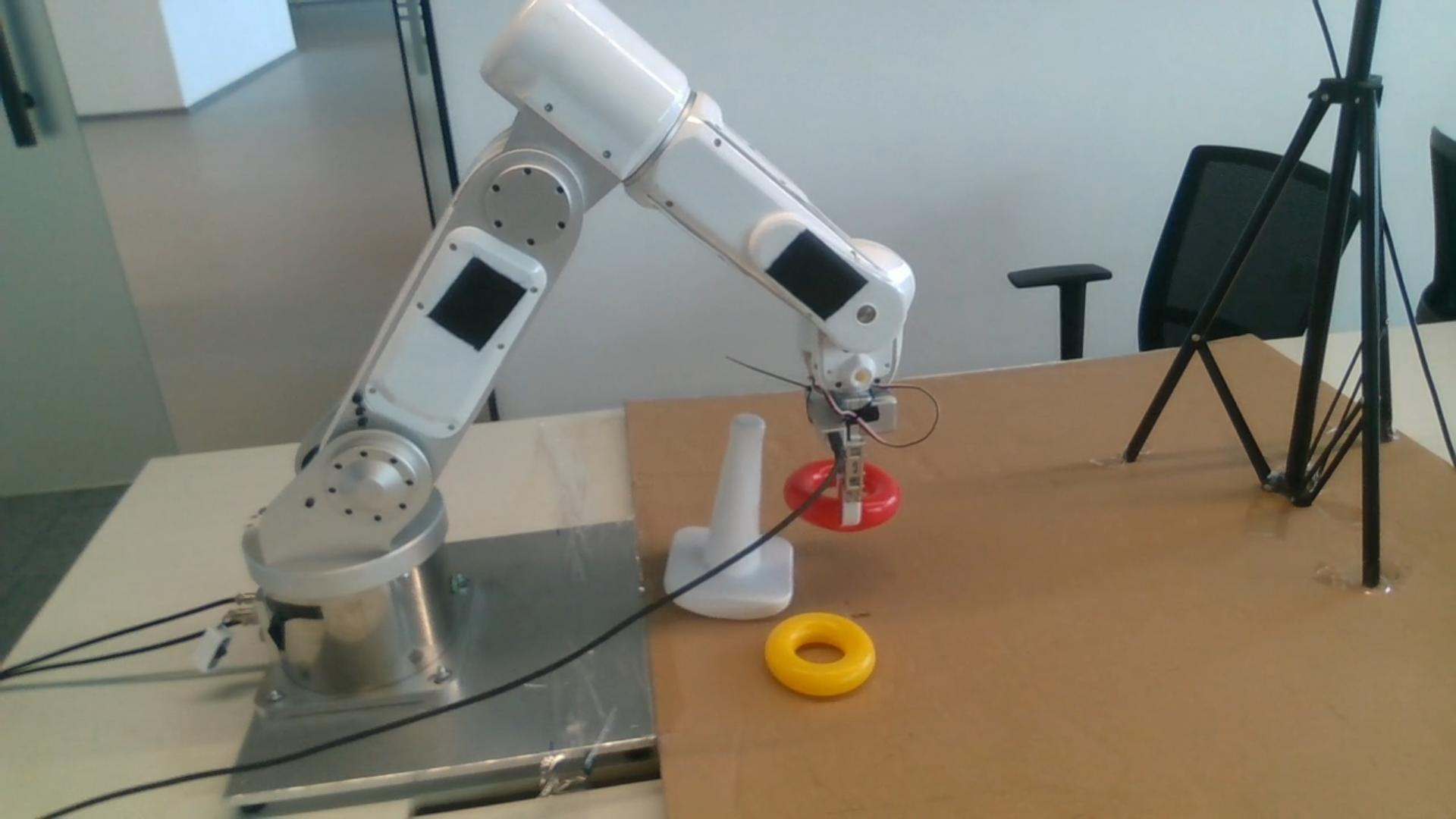}
    \caption*{(d) RingToss}
\end{minipage}
\hfill
\begin{minipage}[b]{0.18\textwidth}
    \centering
    \includegraphics[width=\textwidth]{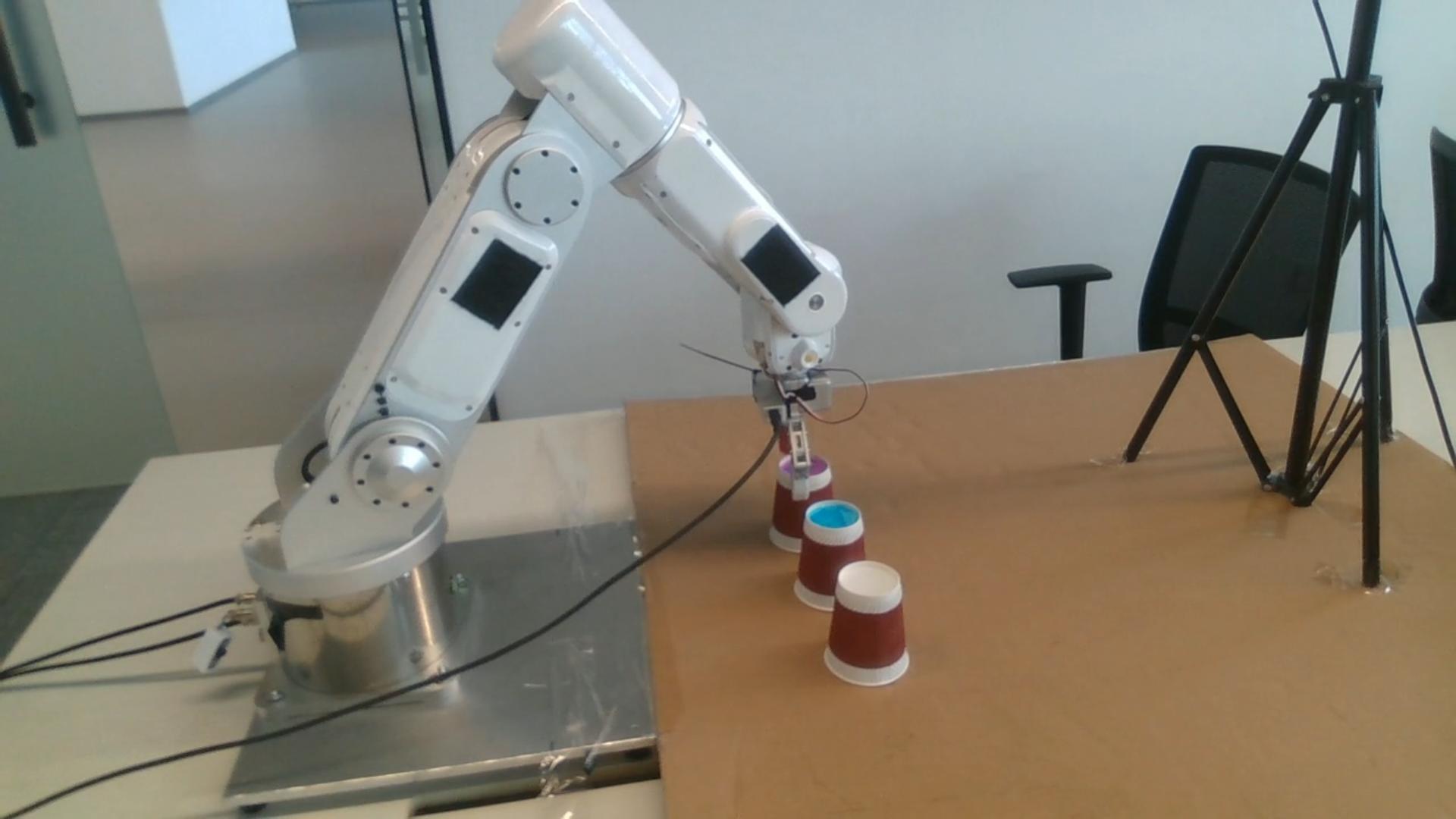}
    \caption*{(e) CupStack}
\end{minipage}

\vspace{0.5cm} 

\begin{minipage}[b]{0.18\textwidth}
    \centering
    \includegraphics[width=\textwidth]{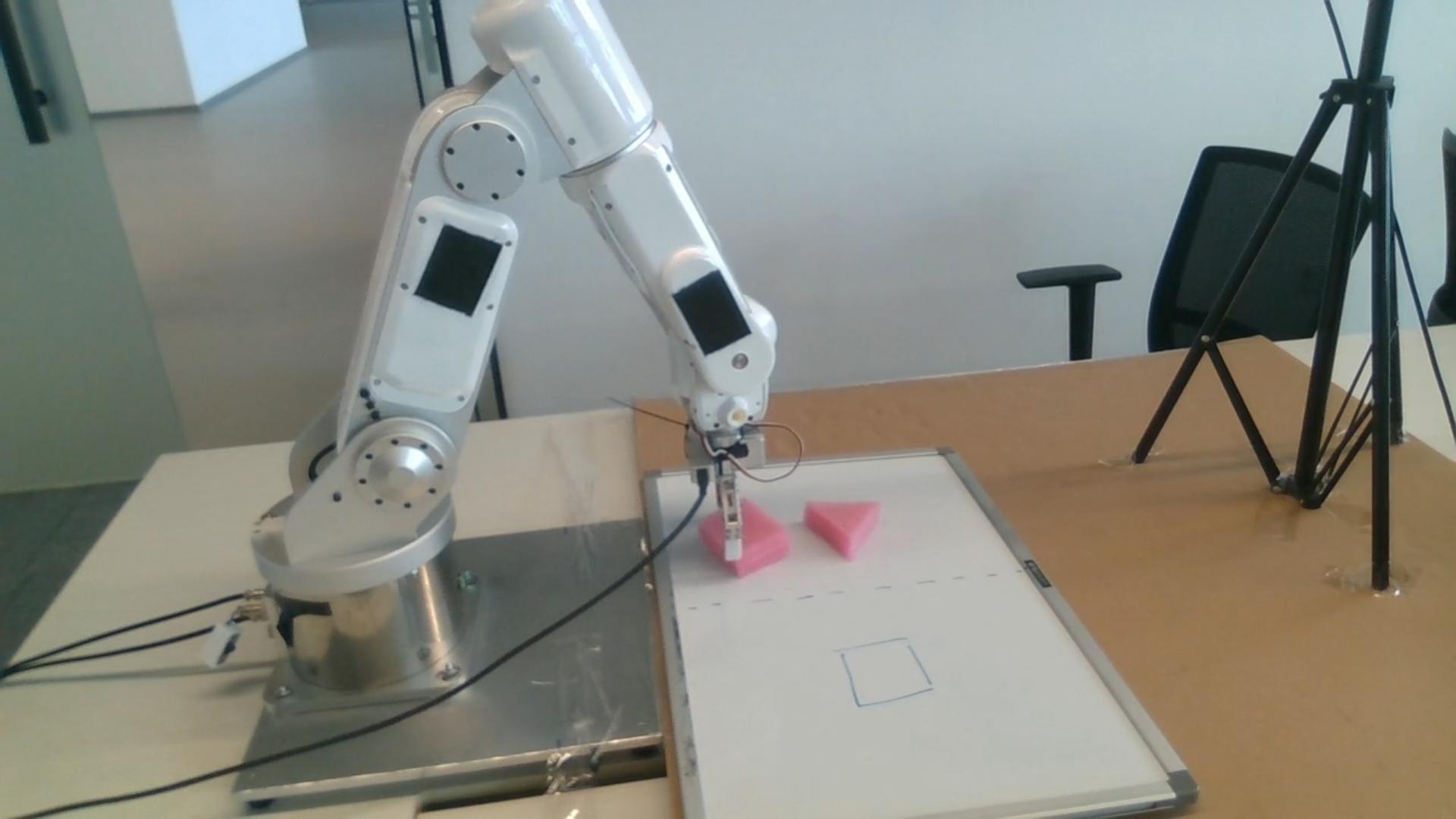}
    \caption*{(f) ShapeDistinguish}
\end{minipage}
\hfill
\begin{minipage}[b]{0.18\textwidth}
    \centering
    \includegraphics[width=\textwidth]{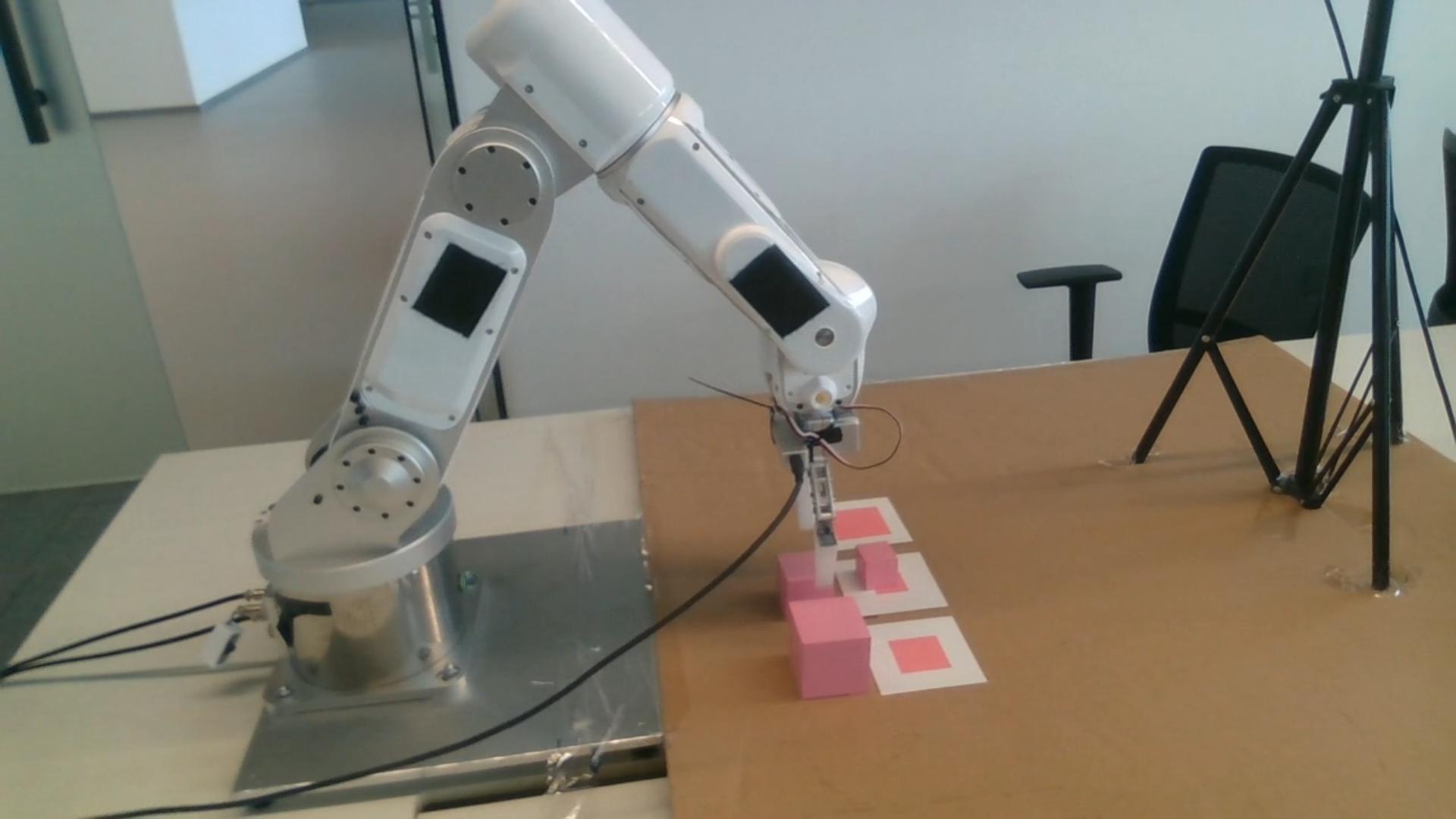}
    \caption*{(g) WhichCube}
\end{minipage}
\hfill
\begin{minipage}[b]{0.18\textwidth}
    \centering
    \includegraphics[width=\textwidth]{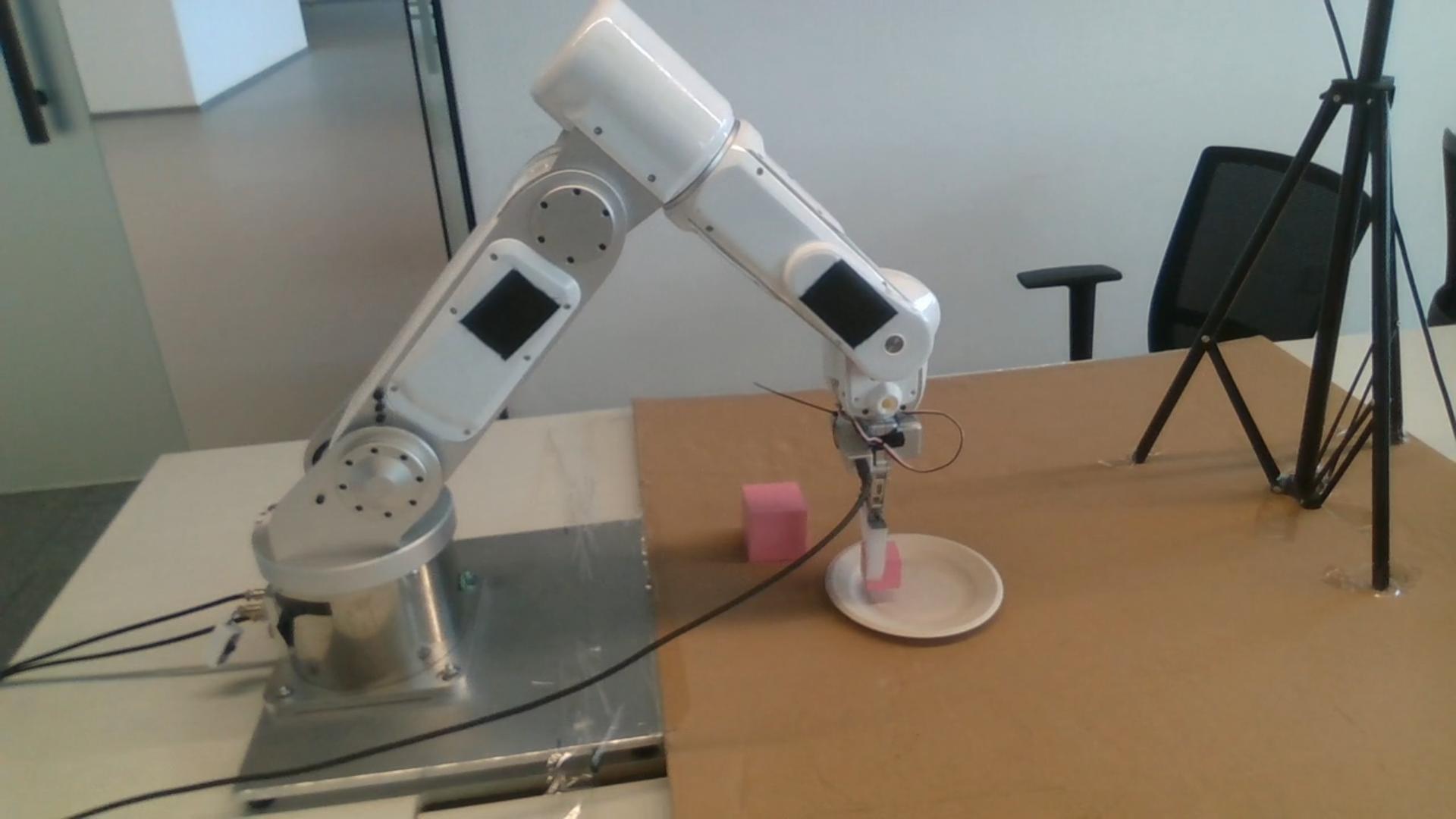}
    \caption*{(h) PickSmall}
\end{minipage}
\hfill
\begin{minipage}[b]{0.18\textwidth}
    \centering
    \includegraphics[width=\textwidth]{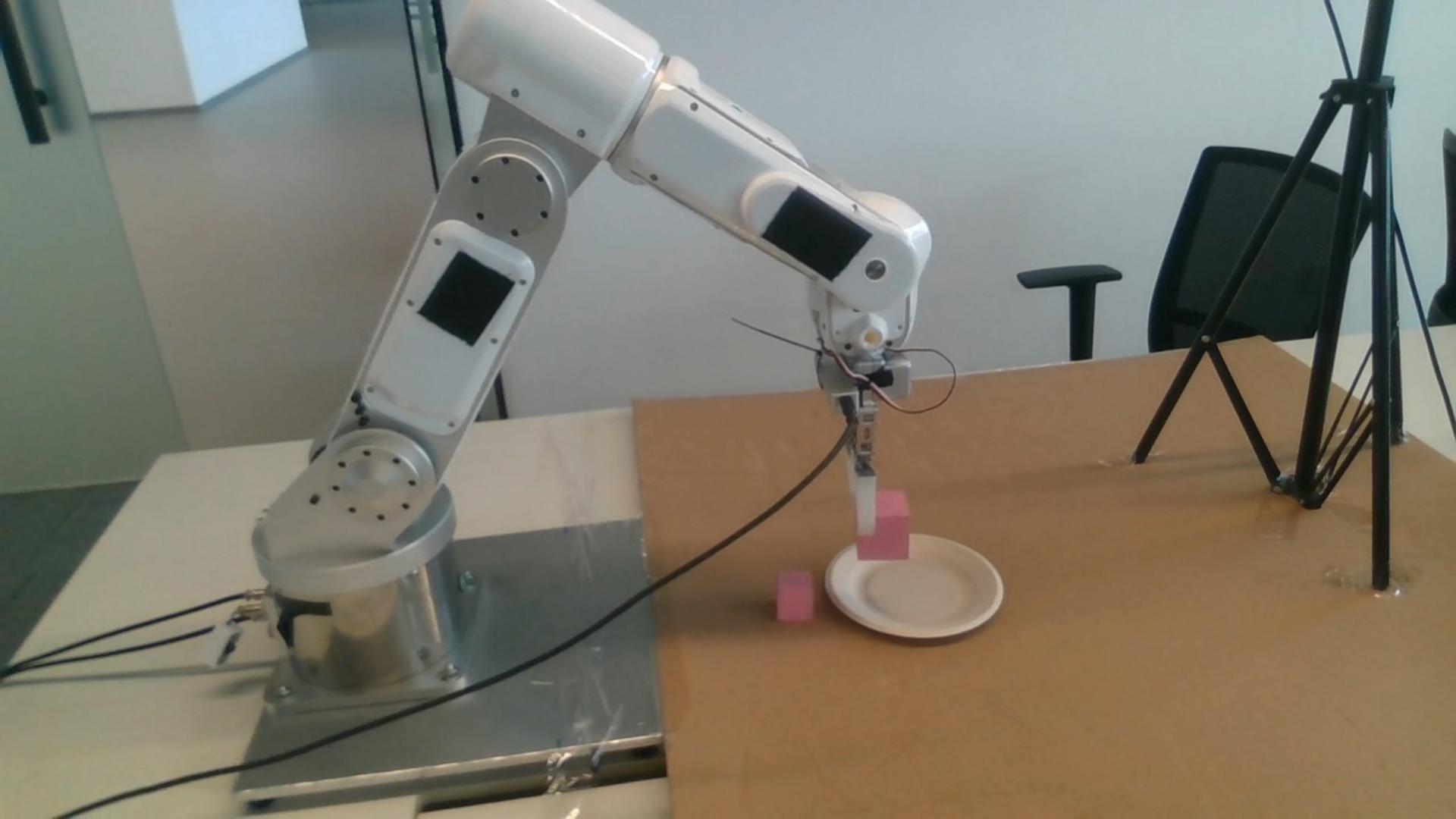}
    \caption*{(i) PickBig}
\end{minipage}
\hfill
\begin{minipage}[b]{0.18\textwidth}
    \centering
    \includegraphics[width=\textwidth]{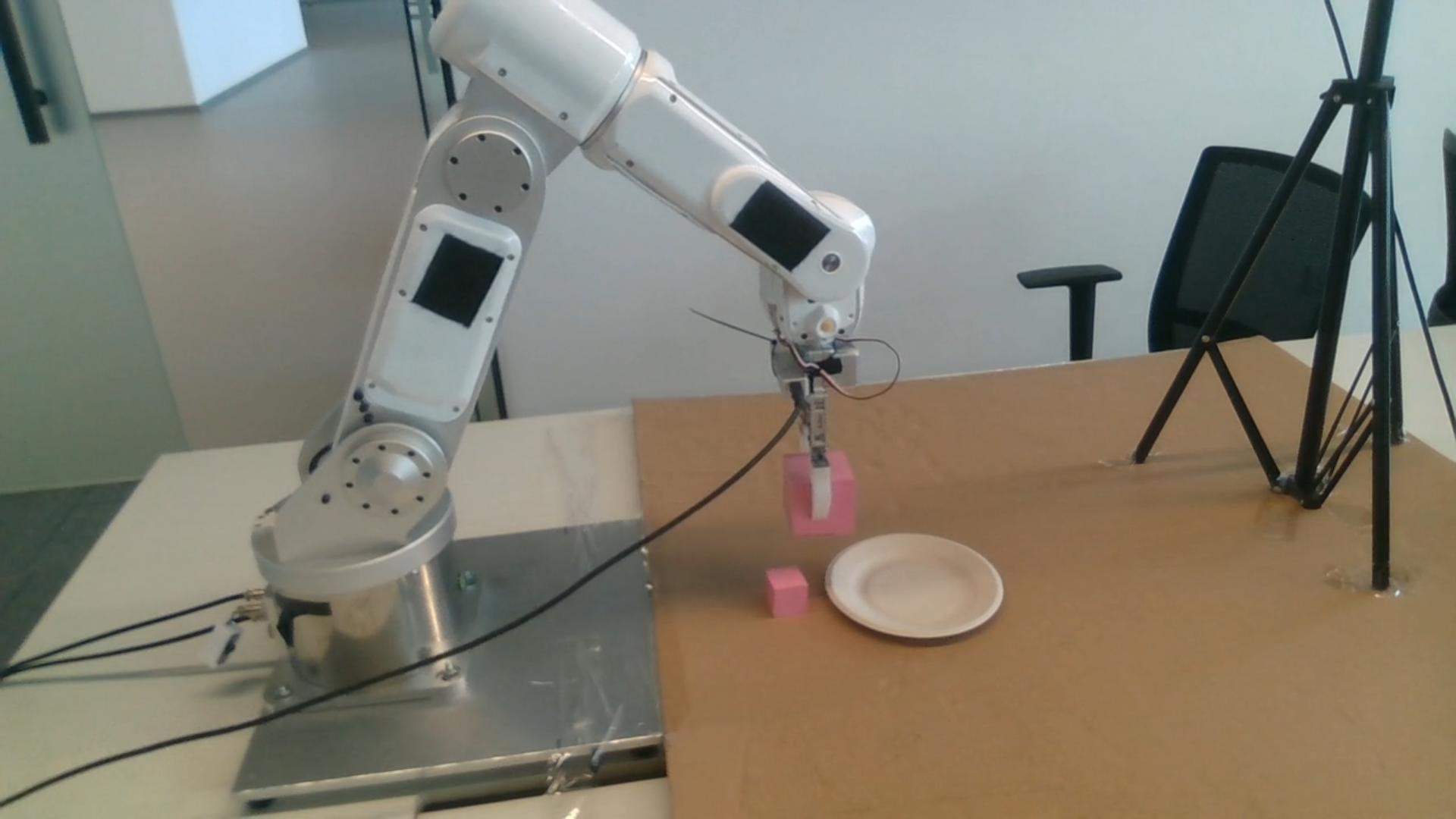}
    \caption*{(j) PickBigV2}
\end{minipage}

\caption{Offline training tasks. For each of the ten training tasks, we collected data for offline training. (a) \textbf{PickPlace}: The robot arm grabs a bottle and places it into a paper plate receptacle. (b) \textbf{BlockPick}: The robot arm picks the red cube to the red plate, then the green cube to the left plate. (c) \textbf{Basketball}: The robot arm picks up a tennis ball and places it into a toy basketball hoop. (d) \textbf{RingToss}: The robot places the red ring onto a white peg, followed by the yellow ring. (e) \textbf{CupStack}: The robot arm stacks the purple cup onto the blue one, then the unpainted cup on top. (f) \textbf{ShapeDistinguish}: The robot selects the correct foam shape based on the drawn shape and transfers it. (g) \textbf{WhichCube}: The robot arm places three pink cubes onto corresponding identity cards. (h) \textbf{PickSmall}: Out of three candidates, the robot arm picks the smallest cubes from two and places them on a plate. (i) \textbf{PickBig}: Out of three candidates, the robot arm picks the largest cubes from two and places them on a plate. (j) \textbf{PickBigV2}: Out of four candidates, the robot arm picks the largest cubes from two and places them on a plate.}
\label{fig:training_tasks}
\end{figure*}

Well-defined tasks are crucial for evaluating model performance and ensuring consistent, quantifiable results. In this study, we present 10 real-world tasks, as shown in Figure \ref{fig:training_tasks}, using low-cost, easily accessible objects to facilitate easy replication. Each task targets specific model capabilities, such as color recognition in "BlockPick" and size differentiation in "PickSmall," allowing us to evaluate the model’s ability to handle diverse features of scenarios in industrial demands.

The tasks are designed to balance simplicity in setup with complexity in execution, providing a broad range of challenges to test the model’s generalization abilities. This approach ensures that the tasks are accessible to the wider research community while offering meaningful insights into the model's performance across different dimensions of perception and decision-making.

\subsection{Voting Positive Rate}
\label{sec:Voting Positive Rate}
Evaluating model performance in real-world environments during training is challenging, as continuous real-time evaluation is often impractical. A common approach is to test models in simulation environments, which can help approximate real-world conditions. However, simulations require meticulous design, including the modeling of physical dynamics and defining success metrics. One key difficulty is determining when a model should be considered to have failed in simulation, a decision that is much easier for humans to make in real-world tests. We chose to forgo simulated evaluation not only because it is difficult to implement with low-cost hardware (designing a custom simulation environment for a personal robot arm may not be feasible), but also because real-world testing is inevitable when transitioning into industry for practical applications. As such, simulated evaluation was deemed unnecessary for the scope of our work.

When it comes to evaluating the model in the real environment, we propose the \textbf{Voting Positive Rate} as a way to mitigate the inherent subjectivity of human judgment. Similar to prior work \cite{c2}, human evaluators assess whether the model successfully completes tasks, but we introduce a voting system to improve reliability. Four evaluators independently judge each task, which is broken down into several cases. Each case is evaluated five times, with initial conditions set by a person who was not involved in the data collection process. A test is deemed successful only if all evaluators vote positively; otherwise, it is classified as a failure. This voting system, referred to as the Voting Positive Rate, helps reduce bias and inconsistency in human evaluations.



\subsection{Checkpoint Selection}
\label{sec:Checkpoint Selection}
Checkpoint selection in imitation learning remains a challenging problem, as optimizing the loss function does not necessarily correlate with achieving the highest success rate in real-world scenarios. As noted in \cite{c2}, the loss curve during training often lacks a clear correlation with actual task performance, which is aligned with the phenomena we also observed in our experiments. Given the impracticality of evaluating model performance after every epoch in real-world tasks, our approach involves training the model overnight, saving checkpoints every 50 epochs, and subsequently evaluating each checkpoint to select the one with the highest success rate.

\subsection{Model Deployment}
The deployment of the trained model is currently executed on our PC, with the generated actions transferred to the robot via a cable. Our alignment strategy involves having the model predict actions along with their corresponding desired timestamps. The robot then moves based on both the action and the timestamp. If a timestamp is deemed invalid (e.g., if it exceeds the allotted time), the action is discarded accordingly.

\section{Experiments and Results}
\label{sec:Experiments and Results}

\subsection{Task Analysis}
Following the standardized checkpoint selection method discussed in Section \ref{sec:Checkpoint Selection}, we present our analysis of the relationship between task success rates and task-specific characteristics. To minimize the impact of model architecture and training strategy, we utilized a consistent architecture and selected the optimal checkpoint based solely on performance, regardless of training duration.

Table \ref{tab:task_performance} summarizes the manual feature extractions for each task, along with corresponding dataset sizes. While it is difficult to establish a direct linear relationship between task features and success rates, our findings suggest the following key perceptions:

\begin{enumerate}
    \item \textbf{Number of Demonstrations}: The number of demonstrations significantly influences the final success rate. This is evident in tasks such as 'PickSmall,' 'PickBig,' and 'PickBigV2,' which serve as controlled experiments where the dataset size is the primary variable. The success rate increases as the dataset size grows, but plateaus at a certain threshold.
    
    \item \textbf{Task Complexity}: Task complexity is directly correlated with the Voting Positive Rate (VPR). As expected, tasks that require more complex decision-making—such as those involving multiple sequential steps—tend to be more challenging. To quantify this, we introduce a manually defined marker, \textbf{Logic Step}, which represents the number of logical deductions a human would need to make to control the robot arm. This concept is inspired by the use of Chain-of-Thought prompting in Large Language Models \cite{c3,c4}..
    
    \item \textbf{Feature Distinguishability}: Based on practical experience, tasks that prominently feature color differentiation appear to benefit more from the ResNet-based perception encoder. This suggests that the current architecture is particularly effective for tasks where visual color features are critical.
\end{enumerate}

\subsection{Model Architecture Ablation Study}



\begin{table*}[htbp]
\centering
\begin{tabular}{lcccccccccrrrrr}
\hline
\multicolumn{1}{c}{\multirow{2}{*}{\textbf{\begin{tabular}[c]{@{}c@{}}Task\\ Name\end{tabular}}}} &
  \multirow{2}{*}{\textbf{\begin{tabular}[c]{@{}c@{}}Object\\ Num\end{tabular}}} &
  \multirow{2}{*}{\textbf{\begin{tabular}[c]{@{}c@{}}Recept\\ Num\end{tabular}}} &
  \multirow{2}{*}{\textbf{Cases}} &
  \multirow{2}{*}{\textbf{Color}} &
  \multirow{2}{*}{\textbf{Size}} &
  \multirow{2}{*}{\textbf{Shape}} &
  \multirow{2}{*}{\textbf{Logic}} &
  \multirow{2}{*}{\textbf{\begin{tabular}[c]{@{}c@{}}Avg\\ Length\end{tabular}}} &
  \multirow{2}{*}{\textbf{\begin{tabular}[c]{@{}c@{}}Demo\\ Num\end{tabular}}} &
  \multicolumn{4}{c}{\textbf{\begin{tabular}[c]{@{}c@{}}Voting Positive Rate\end{tabular}}} \\ \cline{11-14} 
\multicolumn{1}{c}{} &
   &
   &
   &
   &
   &
   &
   &
   &
   &
  \scriptsize\textbf{\begin{tabular}[c]{@{}l@{}}Res18\\ /\\ UNet\end{tabular}} &
  \scriptsize\textbf{\begin{tabular}[c]{@{}l@{}}Res18\\ /\\ TF\end{tabular}} &
  \scriptsize\textbf{\begin{tabular}[c]{@{}l@{}}Res34\\ /\\ TF\end{tabular}} &
  \scriptsize\textbf{\begin{tabular}[c]{@{}l@{}}FPN\\ /\\ TF\end{tabular}} \\ \hline
PickPlace &
  1 &
  1 &
  10 &
  No &
  No &
  No &
  1 &
  16.35s &
  280 &
  \textbf{92\%} &
  \textbf{92\%} &
  80\% &
  82\% \\
BlockPick &
  2 &
  2 &
  10 &
  Yes &
  No &
  No &
  2 &
  24.02s &
  330 &
  63\% &
  \textbf{74\%} &
  56\% &
  70\% \\
Basketball &
  1 &
  1 &
  2 &
  No &
  No &
  No &
  1 &
  22.07s &
  100 &
  92\% &
  \textbf{92\%} &
  80\% &
  86.7\% \\
RingToss &
  2 &
  1 &
  10 &
  Yes &
  No &
  No &
  2 &
  40.28s &
  100 &
  26\% &
  60\% &
  70\% &
  \textbf{72\%} \\
CupStack &
  2 &
  1 &
  10 &
  Yes &
  No &
  No &
  2 &
  28.49s &
  200 &
  26\% &
  57.5\% &
  70\% &
  \textbf{80\%} \\
ShapeDistinguish &
  1 &
  1 &
  10 &
  Yes &
  No &
  Yes &
  2 &
  12.73s &
  120 &
  16.7\% &
  33\% &
  20\% &
  \textbf{42\%} \\
WhichCube &
  3 &
  3 &
  10 &
  No &
  Yes &
  No &
  3 &
  29.69s &
  80 &
  8.3\% &
  33\% &
  \textbf{35\%} &
  \textbf{35\%} \\
PickSmall &
  1 &
  1 &
  6 &
  No &
  Yes &
  No &
  2 &
  11.93s &
  60 &
  16.7\% &
  20\% &
  \textbf{66.7\%} &
  60\% \\
PickBig &
  1 &
  1 &
  6 &
  No &
  Yes &
  No &
  2 &
  11.88s &
  120 &
  26.7\% &
  50\% &
  26.7\% &
  \textbf{76.6\%} \\
PickBigV2 &
  1 &
  1 &
  12 &
  No &
  Yes &
  No &
  2 &
  12.00s &
  240 &
  45\% &
  55.7\% &
  50\% &
  \textbf{78.3\%} \\ \cline{1-14}
\end{tabular}
\caption{\textbf{Overview of task characteristics}: \textbf{Object Num} refers to the number of objects the robot interacts with; \textbf{Recep Num} indicates the number of possible changes to the receptacle; \textbf{Cases} specifies the number of initial conditions; \textbf{Color}, \textbf{Size}, and \textbf{Shape} indicate whether the task involves classifying these features; \textbf{Logic Step} represents the number of logical deductions required; \textbf{Avg Length} is the average duration of each demonstration video; \textbf{Demo Num} denotes the total number of demonstrations for each task. \textbf{Model structure abbreviations:} \textbf{Res18/UNet} refers to ResNet18 with a UNet architecture, \textbf{Res18/TF(Transformer)} refers to ResNet18 with a Transformer-based architecture, \textbf{Res34/TF} refers to ResNet34 with a Transformer-based architecture, and \textbf{FPN/TF} refers to a Feature Pyramid Network (FPN) with a Transformer-based architecture. Each model checkpoint was selected based on exhaustive evaluation mentioned on Section \ref{sec:Checkpoint Selection}.}
\label{tab:task_performance}
\end{table*}
As aforementioned \ref{sec:Policy Control System}, we decoupled the whole controlling system into perception module and action prediction module, and each of them can be instanced by certain network architectures. Hence, we conducted an ablation study to evaluate the performance of various model architectures by modifying both of them. As shown in Table \ref{tab:task_performance}, the study compares the performance of ResNet18 with a CNN based module, and ResNet18, ResNet34, and FPN-based ResNet34 with transformer-based across different tasks.


For simpler tasks such as PickPlace and Basketball, the ResNet18+UNet architecture achieved relatively good VPR, with a notable 92\% rate. This indicates that CNN-based noise prediction networks are effective when the task complexity is relatively low, likely due to their ability to capture local spatial patterns efficiently.

However, transformer-based architectures perform generally better compared to CNN-based. When task complexity increased, the outnum becomes larger. Transformers particularly excelled in the most complex tasks, such as PickSmall, where ResNet34+Transformer outperformed all other architectures with a 66.7\% success rate. This confirms that transformers are well-suited for handling tasks requiring more sophisticated temporal and spatial reasoning, as their self-attention mechanism allows them to capture long-range dependencies more effectively than CNNs.

Among Transformer group, FPN+Transformer is considered as the optimal choice, as it generally perform better. For instance, in the CupStack task, the FPN+Transformer architecture outperformed others, achieving an 80\% success rate, compared to 70\% for ResNet34+Transformer and 57.5\% for ResNet18+Transformer. Similarly, in the PickBig task, FPN+Transformer achieved 76.6\%, significantly outperforming the ResNet18+Transformer (50\%) and ResNet34+Transformer (26.7\%). When simply enlarge the perception model (e.g., from ResNet18 to ResNet34), the improvement of performance is not significant, ResNet34 even perfor worse on some simple tasks (e.g., 'PickPlace', 'BlockPick' and 'Basketball').

The ablation study underscores that while CNN-based architectures may perform well in simpler environments, transformer-based architectures offer a clear advantage in complex, dynamic tasks. These results suggest that, for tasks requiring intricate action sequences and environmental interactions, transformer-based models should be prioritized for their robustness and performance.



\subsection{Dataset Scaling Matters More Than Training Scaling}
\label{sec:Dataset Scaling Matters}
We explicitly explored the effects of scaling both the dataset size and the model architecture by introducing additional demonstrations and increasing the model's hidden dimensions (i.e., adding more learnable parameters) alongside extended training epochs. Interestingly, our results indicate that simply increasing the number of model parameters and training time does not lead to improved performance. In fact, this approach often results in a significant drop in performance, rendering the model's success rate unacceptably low.

Conversely, increasing the number of demonstrations consistently improves model performance, as evidenced by the results from the ‘PickSmall’, ‘PickBig’, and ‘PickBigV2’ tasks. These three tasks were deliberately designed to be similar, with the primary difference being the number of demonstrations. The results are compelling: by increasing the number of demonstrations from 60 to 120, the VPR improves by 9.9\%. However, this improvement tends to plateau; when increasing the demonstrations from 120 to 240, the VPR only improves by 1.7\%. In addition to expanding the dataset, we also experimented with using a larger transformer model, but the results were far from expected. Although we aimed to replicate the scaling laws’ success observed in other domains by simultaneously increasing both dataset size and model complexity, our findings indicate that this strategy requires further exploration to be successfully applied.

\subsection{Data Quality}
\label{sec:Data quality}
During our experiments, we identified a recurring data quality issue, which we present here. When two data collectors independently gathered data for the same tasks, we observed significant variations in model performance, even when the demonstration lengths and training epochs were kept constant. These findings were consistent across most of the tasks we designed, particularly for tasks with longer demonstrations. While it is possible that this performance fluctuation could be attributed to the checkpoint selection problem mentioned earlier, we are inclined to believe it is more closely related to the proficiency level of the data collectors \cite{c1, c2}. However, the definition of “proficiency” remains unclear to us. Therefore, we are simply reporting our perceptions here and leave this question open for further investigation by our peers.

\subsection{Model Generalization}
\label{sec:Model Generalization}
\subsubsection*{Multi-task Generalization}
Similar to the early stages of deep learning, prior work in imitation learning has predominantly focused on training models for single, specific tasks. In contrast, we have developed a multi-task learning approach by combining two datasets and training on a pre-existing checkpoint that was already fine-tuned on one of the tasks. The efficacy of this approach was demonstrated by fine-tuning the ‘BlockPick’ task using a model checkpoint previously trained on the ‘Basketball’ task for 650 epochs. Remarkably, the ‘BlockPick’ task was successfully completed after only 50 additional epochs of fine-tuning, underscoring the efficiency and effectiveness of our multi-task learning strategy.

\subsubsection*{Environmental Scene Generalization}
In standard imitation learning setups, demonstration data is typically collected with fixed camera positions, commonly involving one global view and another camera mounted on the end effector. This setup requires identical camera positions and view angles during inference to replicate the training conditions. Consequently, models trained under these constraints often exhibit limited adaptability and generalization to even minor variations in camera placement in real-world scenarios.
By incorporating merely two distinct camera views, with one camera fixed and the other placed at two different positions (each providing 40 demonstrations), we fine-tuned the model for an additional 100 epochs. Notably, in the 'BlockPick' task, this fine-tuning with multi-angle camera data enabled the model to maintain high performance despite variations in camera positioning, including zero-shot \cite{c53} positions and angles that were entirely unseen during the training phase. This result indicates that the model achieved robust generalization across a range of environmental visual conditions.


\section{Future work}
\label{sec:Discussion and Future work}
Imitation learning should be adaptable across various robotic platforms and industrial tasks, without relying on overly complex or impractical setups. In this study, we utilized a Diffusion Policy-inspired architecture with modifications to the perception module and action prediction module. Future direction could explore the utilization of human language instruction \cite{c43, c45, c46} to strengthen the logic deduction capacity. 

Though our framework can work smoothly from data collection to model deployment, which provides a pipeline for efficient replicating based on data driven paradigm. One of the main obstacle is the required data size, and our demand is to decrease the data size needed while remaining the model performace. One of the convincing method is to leverage the capability of model pretrained on large, open-source robotic trajectory datasets, such as X-Embodiment \cite{c47}, which allows policies to adapt to novel tasks and environments with minimal fine-tuning. 

In summary, our future work will focus on progressively reducing the data dependency of our framework by leveraging advanced transfer learning models and methods. We believe this approach will contribute significantly to the robot learning community.

\section{Conclusion}
In this work, we present a cost-effective and generalized framework for deploying robotic systems in industry-relevant tasks, significantly reducing hardware expenses compared to conventional research setups. Our methodology minimizes the time required for data collection and model training. This efficiency makes the framework highly accessible to a wide range of users, from academic researchers to industry practitioners. 


We also provided detailed guidelines for task design, success rate evaluation, and optimal checkpoint selection. Our experiments demonstrated the feasibility of training multi-task models on real-world tasks, and we observed that even minor adjustments to the model architecture can result in substantial performance improvements across different tasks. These findings contribute valuable insights into optimizing model architectures for diverse, complex environments.

In summary, we introduced a low-cost imitation learning framework supported by a dataset of 10 real-world tasks, designed to accelerate progress in embodied intelligence. By fostering research and open-source collaboration, we aim to enable the development of emergent capabilities in robotics, similar to those observed in large-scale language models, thus driving future advancements in autonomous systems.

\bibliography{References}

\begin{thebibliography}{10}

\bibitem{c1}
Cheng Chi, Zhenjia Xu, Siyuan Feng, Eric Cousineau, Yilun Du, Benjamin Burchfiel, Russ Tedrake, and Shuran Song.
\newblock Diffusion policy: Visuomotor policy learning via action diffusion, 2024.

\bibitem{c2}
Ajay Mandlekar, Danfei Xu, Josiah Wong, Soroush Nasiriany, Chen Wang, Rohun Kulkarni, Li~Fei-Fei, Silvio Savarese, Yuke Zhu, and Roberto Martín-Martín.
\newblock What matters in learning from offline human demonstrations for robot manipulation, 2021.

\bibitem{c45}
Octo~Model Team, Dibya Ghosh, Homer Walke, Karl Pertsch, Kevin Black, Oier Mees, Sudeep Dasari, Joey Hejna, Tobias Kreiman, Charles Xu, Jianlan Luo, You~Liang Tan, Lawrence~Yunliang Chen, Pannag Sanketi, Quan Vuong, Ted Xiao, Dorsa Sadigh, Chelsea Finn, and Sergey Levine.
\newblock Octo: An open-source generalist robot policy, 2024.

\bibitem{c48}
Christopher~G Atkeson and Stefan Schaal.
\newblock Robot learning from demonstration.
\newblock In {\em ICML}, volume~97, pages 12--20, 1997.

\bibitem{c49}
Richard~P Paul.
\newblock {\em Robot manipulators: mathematics, programming, and control: the computer control of robot manipulators}.
\newblock Richard Paul, 1981.

\bibitem{c50}
Josh Achiam, Steven Adler, Sandhini Agarwal, Lama Ahmad, Ilge Akkaya, Florencia~Leoni Aleman, Diogo Almeida, Janko Altenschmidt, Sam Altman, Shyamal Anadkat, et~al.
\newblock Gpt-4 technical report.
\newblock {\em arXiv preprint arXiv:2303.08774}, 2023.

\bibitem{c5}
Tianhao Zhang, Zoe McCarthy, Owen Jow, Dennis Lee, Xi~Chen, Ken Goldberg, and Pieter Abbeel.
\newblock Deep imitation learning for complex manipulation tasks from virtual reality teleoperation.
\newblock In {\em 2018 IEEE International Conference on Robotics and Automation (ICRA)}, pages 5628--5635, 2018.

\bibitem{c6}
Ajay Mandlekar, Danfei Xu, Roberto Mart{\'\i}n-Mart{\'\i}n, Silvio Savarese, and Li~Fei-Fei.
\newblock Learning to generalize across long-horizon tasks from human demonstrations.
\newblock {\em arXiv preprint arXiv:2003.06085}, 2020.

\bibitem{c7}
Agrim Gupta, Silvio Savarese, Surya Ganguli, and Li~Fei-Fei.
\newblock Embodied intelligence via learning and evolution.
\newblock {\em Nature communications}, 12(1):5721, 2021.

\bibitem{c8}
Heecheol Kim, Yoshiyuki Ohmura, and Yasuo Kuniyoshi.
\newblock Robot peels banana with goal-conditioned dual-action deep imitation learning.
\newblock {\em arXiv preprint arXiv:2203.09749}, 2022.

\bibitem{c9}
Oliver Kroemer, Christian Daniel, Gerhard Neumann, Herke Van~Hoof, and Jan Peters.
\newblock Towards learning hierarchical skills for multi-phase manipulation tasks.
\newblock In {\em 2015 IEEE international conference on robotics and automation (ICRA)}, pages 1503--1510. IEEE, 2015.

\bibitem{c25}
Dean~A. Pomerleau.
\newblock Efficient training of artificial neural networks for autonomous navigation.
\newblock {\em Neural Computation}, 3(1):88--97, 1991.

\bibitem{c26}
Stephane Ross, Geoffrey~J. Gordon, and J.~Andrew Bagnell.
\newblock A reduction of imitation learning and structured prediction to no-regret online learning, 2011.

\bibitem{c13}
Tony~Z Zhao, Vikash Kumar, Sergey Levine, and Chelsea Finn.
\newblock Learning fine-grained bimanual manipulation with low-cost hardware.
\newblock {\em arXiv preprint arXiv:2304.13705}, 2023.

\bibitem{c24}
Pete Florence, Corey Lynch, Andy Zeng, Oscar Ramirez, Ayzaan Wahid, Laura Downs, Adrian Wong, Johnny Lee, Igor Mordatch, and Jonathan Tompson.
\newblock Implicit behavioral cloning, 2021.

\bibitem{c27}
Daniel Jarrett, Ioana Bica, and Mihaela van~der Schaar.
\newblock Strictly batch imitation learning by energy-based distribution matching, 2021.

\bibitem{c51}
Michael Janner, Yilun Du, Joshua~B. Tenenbaum, and Sergey Levine.
\newblock Planning with diffusion for flexible behavior synthesis, 2022.

\bibitem{c52}
Jonathan Ho, Ajay Jain, and Pieter Abbeel.
\newblock Denoising diffusion probabilistic models.
\newblock {\em Advances in neural information processing systems}, 33:6840--6851, 2020.

\bibitem{c28}
Zoltán Lőrincz, Márton Szemenyei, and Róbert Moni.
\newblock Imitation learning for generalizable self-driving policy with sim-to-real transfer, 2022.

\bibitem{c29}
Wenshuai Zhao, Jorge~Pena Queralta, Li~Qingqing, and Tomi Westerlund.
\newblock Towards closing the sim-to-real gap in collaborative multi-robot deep reinforcement learning.
\newblock In {\em 2020 5th International conference on robotics and automation engineering (ICRAE)}, pages 7--12. IEEE, 2020.

\bibitem{c31}
Marcel Torne, Anthony Simeonov, Zechu Li, April Chan, Tao Chen, Abhishek Gupta, and Pulkit Agrawal.
\newblock Reconciling reality through simulation: A real-to-sim-to-real approach for robust manipulation, 2024.

\bibitem{c32}
Shital Shah, Debadeepta Dey, Chris Lovett, and Ashish Kapoor.
\newblock Airsim: High-fidelity visual and physical simulation for autonomous vehicles.
\newblock In {\em Field and Service Robotics}, 2017.

\bibitem{c33}
Alexey Dosovitskiy, German Ros, Felipe Codevilla, Antonio Lopez, and Vladlen Koltun.
\newblock {CARLA}: {An} open urban driving simulator.
\newblock In {\em Proceedings of the 1st Annual Conference on Robot Learning}, pages 1--16, 2017.

\bibitem{c34}
Fadri Furrer, Michael Burri, Markus Achtelik, and Roland Siegwart.
\newblock {\em Robot Operating System (ROS): The Complete Reference (Volume 1)}, chapter RotorS---A Modular Gazebo MAV Simulator Framework, pages 595--625.
\newblock Springer International Publishing, Cham, 2016.

\bibitem{c30}
Ramya Ramakrishnan, Ece Kamar, Debadeepta Dey, Eric Horvitz, and Julie Shah.
\newblock Blind spot detection for safe sim-to-real transfer.
\newblock {\em Journal of Artificial Intelligence Research}, 67:191--234, 02 2020.

\bibitem{c35}
Nur Muhammad~Mahi Shafiullah, Anant Rai, Haritheja Etukuru, Yiqian Liu, Ishan Misra, Soumith Chintala, and Lerrel Pinto.
\newblock On bringing robots home.
\newblock {\em arXiv preprint arXiv:2311.16098}, 2023.

\bibitem{c36}
Cheng Chi, Zhenjia Xu, Chuer Pan, Eric Cousineau, Benjamin Burchfiel, Siyuan Feng, Russ Tedrake, and Shuran Song.
\newblock Universal manipulation interface: In-the-wild robot teaching without in-the-wild robots, 2024.

\bibitem{c40}
Kaiming He, Xiangyu Zhang, Shaoqing Ren, and Jian Sun.
\newblock Deep residual learning for image recognition.
\newblock In {\em Proceedings of the IEEE conference on computer vision and pattern recognition}, pages 770--778, 2016.

\bibitem{c41}
Tsung-Yi Lin, Piotr Doll{\'a}r, Ross Girshick, Kaiming He, Bharath Hariharan, and Serge Belongie.
\newblock Feature pyramid networks for object detection.
\newblock In {\em Proceedings of the IEEE conference on computer vision and pattern recognition}, pages 2117--2125, 2017.

\bibitem{c3}
Takeshi Kojima, Shixiang~(Shane) Gu, Machel Reid, Yutaka Matsuo, and Yusuke Iwasawa.
\newblock Large language models are zero-shot reasoners.
\newblock In S.~Koyejo, S.~Mohamed, A.~Agarwal, D.~Belgrave, K.~Cho, and A.~Oh, editors, {\em Advances in Neural Information Processing Systems}, volume~35, pages 22199--22213. Curran Associates, Inc., 2022.

\bibitem{c4}
Jason Wei, Xuezhi Wang, Dale Schuurmans, Maarten Bosma, brian ichter, Fei Xia, Ed~Chi, Quoc~V Le, and Denny Zhou.
\newblock Chain-of-thought prompting elicits reasoning in large language models.
\newblock In S.~Koyejo, S.~Mohamed, A.~Agarwal, D.~Belgrave, K.~Cho, and A.~Oh, editors, {\em Advances in Neural Information Processing Systems}, volume~35, pages 24824--24837. Curran Associates, Inc., 2022.

\bibitem{c53}
Tom~B Brown.
\newblock Language models are few-shot learners.
\newblock {\em arXiv preprint arXiv:2005.14165}, 2020.

\bibitem{c43}
Siddharth Karamcheti, Suraj Nair, Annie~S Chen, Thomas Kollar, Chelsea Finn, Dorsa Sadigh, and Percy Liang.
\newblock Language-driven representation learning for robotics.
\newblock {\em arXiv preprint arXiv:2302.12766}, 2023.

\bibitem{c46}
Ria Doshi, Homer Walke, Oier Mees, Sudeep Dasari, and Sergey Levine.
\newblock Scaling cross-embodied learning: One policy for manipulation, navigation, locomotion and aviation.
\newblock {\em arXiv preprint arXiv:2408.11812}, 2024.

\bibitem{c47}
Embodiment Collaboration, Abby O'Neill, Abdul Rehman, Abhinav Gupta, Abhiram Maddukuri, Abhishek Gupta, Abhishek Padalkar, Abraham Lee, Acorn Pooley, Agrim Gupta, Ajay Mandlekar, Ajinkya Jain, Albert Tung, Alex Bewley, Alex Herzog, Alex Irpan, Alexander Khazatsky, Anant Rai, Anchit Gupta, Andrew Wang, Andrey Kolobov, Anikait Singh, Animesh Garg, Aniruddha Kembhavi, Annie Xie, Anthony Brohan, Antonin Raffin, Archit Sharma, Arefeh Yavary, Arhan Jain, Ashwin Balakrishna, Ayzaan Wahid, Ben Burgess-Limerick, Beomjoon Kim, Bernhard Schölkopf, Blake Wulfe, Brian Ichter, Cewu Lu, Charles Xu, Charlotte Le, Chelsea Finn, Chen Wang, Chenfeng Xu, Cheng Chi, Chenguang Huang, Christine Chan, Christopher Agia, Chuer Pan, Chuyuan Fu, Coline Devin, Danfei Xu, Daniel Morton, Danny Driess, Daphne Chen, Deepak Pathak, Dhruv Shah, Dieter Büchler, Dinesh Jayaraman, Dmitry Kalashnikov, Dorsa Sadigh, Edward Johns, Ethan Foster, Fangchen Liu, Federico Ceola, Fei Xia, Feiyu Zhao, Felipe~Vieira Frujeri, Freek Stulp, Gaoyue Zhou,
  Gaurav~S. Sukhatme, Gautam Salhotra, Ge~Yan, Gilbert Feng, Giulio Schiavi, Glen Berseth, Gregory Kahn, Guangwen Yang, Guanzhi Wang, Hao Su, Hao-Shu Fang, Haochen Shi, Henghui Bao, Heni~Ben Amor, Henrik~I Christensen, Hiroki Furuta, Homanga Bharadhwaj, Homer Walke, Hongjie Fang, Huy Ha, Igor Mordatch, Ilija Radosavovic, Isabel Leal, Jacky Liang, Jad Abou-Chakra, Jaehyung Kim, Jaimyn Drake, Jan Peters, Jan Schneider, Jasmine Hsu, Jay Vakil, Jeannette Bohg, Jeffrey Bingham, Jeffrey Wu, Jensen Gao, Jiaheng Hu, Jiajun Wu, Jialin Wu, Jiankai Sun, Jianlan Luo, Jiayuan Gu, Jie Tan, Jihoon Oh, Jimmy Wu, Jingpei Lu, Jingyun Yang, Jitendra Malik, João Silvério, Joey Hejna, Jonathan Booher, Jonathan Tompson, Jonathan Yang, Jordi Salvador, Joseph~J. Lim, Junhyek Han, Kaiyuan Wang, Kanishka Rao, Karl Pertsch, Karol Hausman, Keegan Go, Keerthana Gopalakrishnan, Ken Goldberg, Kendra Byrne, Kenneth Oslund, Kento Kawaharazuka, Kevin Black, Kevin Lin, Kevin Zhang, Kiana Ehsani, Kiran Lekkala, Kirsty Ellis, Krishan Rana,
  Krishnan Srinivasan, Kuan Fang, Kunal~Pratap Singh, Kuo-Hao Zeng, Kyle Hatch, Kyle Hsu, Laurent Itti, Lawrence~Yunliang Chen, Lerrel Pinto, Li~Fei-Fei, Liam Tan, Linxi~"Jim" Fan, Lionel Ott, Lisa Lee, Luca Weihs, Magnum Chen, Marion Lepert, Marius Memmel, Masayoshi Tomizuka, Masha Itkina, Mateo~Guaman Castro, Max Spero, Maximilian Du, Michael Ahn, Michael~C. Yip, Mingtong Zhang, Mingyu Ding, Minho Heo, Mohan~Kumar Srirama, Mohit Sharma, Moo~Jin Kim, Naoaki Kanazawa, Nicklas Hansen, Nicolas Heess, Nikhil~J Joshi, Niko Suenderhauf, Ning Liu, Norman~Di Palo, Nur Muhammad~Mahi Shafiullah, Oier Mees, Oliver Kroemer, Osbert Bastani, Pannag~R Sanketi, Patrick~"Tree" Miller, Patrick Yin, Paul Wohlhart, Peng Xu, Peter~David Fagan, Peter Mitrano, Pierre Sermanet, Pieter Abbeel, Priya Sundaresan, Qiuyu Chen, Quan Vuong, Rafael Rafailov, Ran Tian, Ria Doshi, Roberto Mart'in-Mart'in, Rohan Baijal, Rosario Scalise, Rose Hendrix, Roy Lin, Runjia Qian, Ruohan Zhang, Russell Mendonca, Rutav Shah, Ryan Hoque, Ryan Julian,
  Samuel Bustamante, Sean Kirmani, Sergey Levine, Shan Lin, Sherry Moore, Shikhar Bahl, Shivin Dass, Shubham Sonawani, Shubham Tulsiani, Shuran Song, Sichun Xu, Siddhant Haldar, Siddharth Karamcheti, Simeon Adebola, Simon Guist, Soroush Nasiriany, Stefan Schaal, Stefan Welker, Stephen Tian, Subramanian Ramamoorthy, Sudeep Dasari, Suneel Belkhale, Sungjae Park, Suraj Nair, Suvir Mirchandani, Takayuki Osa, Tanmay Gupta, Tatsuya Harada, Tatsuya Matsushima, Ted Xiao, Thomas Kollar, Tianhe Yu, Tianli Ding, Todor Davchev, Tony~Z. Zhao, Travis Armstrong, Trevor Darrell, Trinity Chung, Vidhi Jain, Vikash Kumar, Vincent Vanhoucke, Wei Zhan, Wenxuan Zhou, Wolfram Burgard, Xi~Chen, Xiangyu Chen, Xiaolong Wang, Xinghao Zhu, Xinyang Geng, Xiyuan Liu, Xu~Liangwei, Xuanlin Li, Yansong Pang, Yao Lu, Yecheng~Jason Ma, Yejin Kim, Yevgen Chebotar, Yifan Zhou, Yifeng Zhu, Yilin Wu, Ying Xu, Yixuan Wang, Yonatan Bisk, Yongqiang Dou, Yoonyoung Cho, Youngwoon Lee, Yuchen Cui, Yue Cao, Yueh-Hua Wu, Yujin Tang, Yuke Zhu, Yunchu
  Zhang, Yunfan Jiang, Yunshuang Li, Yunzhu Li, Yusuke Iwasawa, Yutaka Matsuo, Zehan Ma, Zhuo Xu, Zichen~Jeff Cui, Zichen Zhang, Zipeng Fu, and Zipeng Lin.
\newblock Open x-embodiment: Robotic learning datasets and rt-x models, 2024.

\end{thebibliography}

\end{document}